\documentclass[11pt,letterpaper]{article}
\usepackage{eacl2017}
\usepackage{times}
\usepackage{latexsym}

\eaclfinalcopy

\usepackage{url}
\usepackage{multirow}
\usepackage{times}
\usepackage{url}
\usepackage{rtrees}
\usepackage{pstricks}
\usepackage{amsmath}
\usepackage{amsfonts}
\usepackage{amssymb}
\usepackage{graphicx}
\usepackage[caption=false,font=footnotesize]{subfig}
\usepackage{cleveref}
\usepackage[T1]{fontenc}

\newcommand{\ignore}[1]{}

\newcommand{\memnn}[0]{\textsc{MemN2N }}

\newcommand{\memnns}[0]{\textsc{MemN2N}s }

\usepackage{url}
\usepackage{dsttr}
\begin{document}

%


\title{Challenging Neural Dialogue Models with Natural Data: Memory Networks Fail on Incremental Phenomena}



 

\author{ 
   Igor Shalyminov\\
 Interaction Lab   \\
 Heriot-Watt University  \\
 {\tt is33@hw.ac.uk} \\
\And
Arash Eshghi \\
  Interaction Lab   \\
 Heriot-Watt University  \\
  {\tt  a.eshghi@hw.ac.uk} \\ \And
  Oliver Lemon \\
 Interaction Lab   \\
 Heriot-Watt University  \\
  {\tt o.lemon@hw.ac.uk}}

\maketitle

\begin{abstract}
		Natural, spontaneous dialogue proceeds incrementally on a word-by-word basis; and it contains many sorts of disfluency such as mid-utterance/sentence hesitations, interruptions, and self-corrections. But training data for machine learning approaches to dialogue processing is often either cleaned-up or wholly synthetic in order to avoid such phenomena. The question then arises of how well systems trained on such clean data generalise to real spontaneous dialogue, or indeed whether they are trainable at all on naturally occurring dialogue data. To answer this question, we created a new corpus called bAbI+\footnote{this   dataset is freely available at \url{https://bit.ly/babi_plus}} by systematically adding natural spontaneous incremental dialogue phenomena such as \emph{restart}s and \emph{self-correction}s to the Facebook AI Research's bAbI dialogues dataset. We then explore the performance of a state-of-the-art retrieval model, MemN2N \cite{babi,memn2n}, on this more natural dataset. Results show that the semantic accuracy of the MemN2N model drops drastically; and that although it is in principle able to learn to process the constructions in bAbI+, it needs an impractical amount of training data to do so. Finally, we go on to show that an incremental, semantic parser~-- DyLan~-- shows 100\% semantic accuracy on both bAbI and bAbI+, highlighting the generalisation properties of linguistically informed dialogue models.

\end{abstract}



\section{Introduction}\label{tab:introduction}

A key problem for the practical data-driven  (rather than hand-crafted) development of task-oriented dialogue systems is that   they are generally turn-based, and so do not support natural, everyday  {\it incremental} (i.e.\ word-by-word) dialogue processing. This means that they often cannot process naturally occurring incremental dialogue phenomena such as mid-sentence restarts and self-corrections \cite{Hough15,Howes.etal09}. Dialogue systems will not be able to make sense of the everyday language produced by users which is replete with pauses, interruptions, self-corrections and other inherently incremental dialogue phenomena, until they incorporate one or another form of incremental language processing. Indeed incremental dialogue systems (i.e.\ processing word-by-word instead of at utterance/turn boundaries) have previously been empirically shown to be beneficial and more natural for users \cite{aistincremental,Skantze.Hjalmarsson10}.


In this paper, we explore the performance of the state-of-the-art neural retrieval model of \newcite{babi} on dialogues containing some prototypical incremental dialogue structures. \newcite{babi} initially presented the bAbI dialog tasks dataset aimed at learning goal-oriented dialogue systems in an end-to-end fashion: there are no annotations in the data whatsoever, and the model learns all components of a dialogue system. On this dataset, they report that End-to-End Memory Networks (henceforth \textsc{MemN2N}s) achieve an impressive 100\% performance on a test set of 1000 dialogues, after being trained on 1000 similar dialogues.

However, the bAbI dataset is both synthetic and clean: it contains none of the more interesting naturally occurring, disfluent phenomena identified above. To assess the effectiveness of the \memnn model on more natural dialogue data, we introduce an extended, incremental version of the bAbI dataset~-- dubbed bAbI+ (see section \ref{babi+})~-- which we created by systematically adding self-corrections, hesitations, and restarts to the original bAbI dataset.

We go on to explore the performance of \memnn on this new dataset. The results of our experiments show that the semantic accuracy of \memnn, measured in terms of how well the model predicts API calls (a non-linguistic action~--  in this case querying a data-base with the user's requirements) at the end of a dialogue segment, drops very significantly (by about 50\%) even when trained on the full bAbI+ dataset.




Finally, we compare these results to the methodologically distinct, linguistically informed model of \cite{Eshghi.etal17,Kalatzis.etal16}, who employ an incremental dialogue parser, \texttt{DyLan} \cite{Eshghi15,Eshghi.etal11,Purver.etal11}; based around the Dynamic Syntax grammar framework \cite{Kempson.etal01,Cann.etal05a}). We show here that there is no drop in performance in the same semantic accuracy metric from bAbI to bAbI+ with both at 100\% due to the rich, theoretically-grounded knowledge incorporated within the model.

\section{Exploring the performance of \memnns}\label{approach}
%
%
  Our focus in this paper is  to explore the approach of Bordes et al. \shortcite{babi}, and its performance on spontaneous dialogue data.

\subsection{The Dialog bAbI tasks dataset}\label{babi}
We use  Facebook AI Research's Dialogue bAbI tasks dataset \cite{babi}. These are goal-oriented dialogues in the domain of restaurant search. In the dataset, there are 6 tasks of increasing complexity ranging from only collecting the user's preferences on restaurant and up to conducting full dialogues with changes in the user's goal and providing extra information upon request. The first 5 tasks are `clean' dialogues composed synthetically and they thus lack the features of natural everyday conversations. Task 6, in turn, is based on real dialogues collected for the Dialog State Tracking Challenge 2.

\begin{figure*}[ht]
\includegraphics[width=\linewidth]{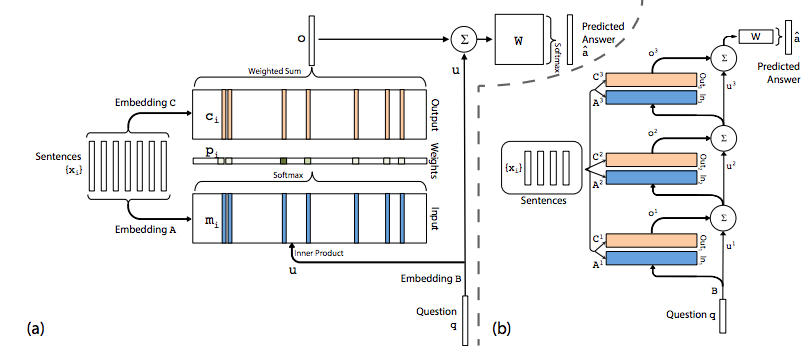}
\caption{MemN2N architecture~-- (a) single memory hop and (b) multiple memory hops, from \cite{memn2n} }\label{fig:memn2n} 
\end{figure*}

Recent studies have shown different ways in which \memnns are outperformed: \newcite{seq2seq_babi} introduced the Copy-Augmented Sequence-to-Sequence model that outperforms \memnn on Task 6; \newcite{hybrid-code-networks} presented a hybrid RNN + rule-based model trainable in a 2-stage supervised + reinforcement learning setup, outperforming \memnn on Tasks 5 and 6.

However, none of these studies control for {\it the type of complexity} that might result in worse performance, and thus do not shed any light on why a particular architecture such as \memnn might be at a disadvantage. While Task 5 dialogues have the full task complexity, conducting full dialogues with an unfixed user goal and additional information requests, they are still composed programmatically and contain minimal surface variation. The Task 6 dialogues on the other hand are complex both in terms of the surface variation and the task itself.

Here, in order to study the specific effects of  incremental variations in dialogue such as  conversational disfluencies, we focus on   Task 1, where in each dialogue the system asks the user about their preferences for the properties of a restaurant, and each dialogue results in an {\it API call} containing values of each slot obtained (e.g.\ {\tt food-type=french})~-- the ability of predicting the API calls correctly thus provides a direct measure of how a well a particular model can interpret the dialogues. 

Using the \memnn model, the approach of \newcite{babi} achieves 100\% performance~-- measured as per-utterance accuracy including the final API call~-- after training on 1000 dialogues.

\subsection{The bAbI+ dataset}\label{babi+}
While containing sufficient lexical variation, the original bAbI Task 1 dialogues significantly lack incremental and interactional variations vital for natural real-life dialogues. In order to obtain such variation while keeping the controllable environment close to the laboratory conditions that bAbI offers, we created the bAbI+ dataset by systematically transforming the original dataset's dialogues. 


bAbI+ is an  extension of the bAbI Task 1 dialogues with everyday incremental dialogue phenomena (hesitations, restarts, and corrections~-- see below). This extension can be seen as orthogonal to the increasing task complexity which Tasks 2--6 offer: we instead increase the complexity of surface forms of dialogue utterances, while keeping every other aspect of the task fixed.

Our incremental modifications model the disfluencies and communication problems in everyday spoken interaction in  real-world environments.
These variations are:
\begin{itemize}

\item \textbf{1. Hesitations}, e.g.\ as in ``we will be \texttt{uhm} eight''; 

\item \textbf{2. Restarts}, e.g.\ ``can you make a restaurant \texttt{uhm yeah can you make a restaurant} reservation for four people with french cuisine in a moderate price range'';

 \item \textbf{3. Corrections} affecting task-specific information~-- both short-distance ones correcting one token, e.g.\ ``with french \texttt{oh no spanish} food'', and long-distance NP/PP-level corrections, e.g. ``with french food \texttt{uhm sorry with spanish food}'', all within a single user utterance, rather than across multiple turns.

\end{itemize}

The phenomena above are mixed in probabilistically from the fixed sets of templates to the original data\footnote{See \url{https://github.com/ishalyminov/babi_tools}}. The  modifications affect a total of \textbf{11336} utterances in the \textbf{3998} dialogues. Around \textbf{21\%} of user's turns contain corrections, \textbf{40\%}~ hesitations, and  \textbf{5\%} restarts (they are not mutually exclusive, so that an utterance can contain up to 3 modifications). Our modifications, with respect to corrections in particular, are more conservative than those observed in real-world data: \cite{Hough15} reports that self-corrections appear in \textbf{20\%} of all turns natural conversations from British National Corpus, and in \textbf{40\%} of turns in the Map Task, a corpus of human-human goal-oriented dialogues.

 Here's part of an example dialogue in the bAbI+ corpus, showing corrections in all user turns:\\

\begin{footnotesize}
\begin{tabular}{lp{6cm}}
\textbf{sys:}&	hello what can I help you with today?\\
\textbf{usr:}&	I'd like to book a uhm yeah I'd like to book a table in a expensive price range no sorry in a cheap price range\\
\textbf{sys:}&	I'm on it. Any preference on a type of cuisine?\\
\textbf{usr:}&	with indian food no sorry with spanish food\\
\textbf{sys:}&	where should it be?\\
\textbf{usr:} &	in tokyo\\
\textbf{sys:}&	how many people would be in your party?\\
\textbf{usr:}&	for uhm yeah for eight people please\\
\textbf{sys:}&	ok let me look into some options for you\\
\end{tabular}
\end{footnotesize}

\subsection{Memory Networks}
We follow Bordes et al.'s setup  by using a MemN2N (we took an open source Tensorflow implementation for bAbI QA tasks and modified it\footnote{See 
\url{https://github.com/ishalyminov/memn2n}
} according to their setup~-- see details below).

The architecture of a MemN2N is shown in Figure~\ref{fig:memn2n} \cite{memn2n}.

One of the model's key features is what can be thought of as  an ``explicit memory'' component: before the inference process, all the input sentences are written into the memory from which they are then read during the inference. The internal memory representation is as follows: an utterance $x_i$ is first vectorized as a fixed-sized array of 1-hot vectors (zero padding is used), then each vector is embedded via the matrix $A$, and finally these embeddings are encoded into a single memory vector $m_i$ using temporal encoding (it preserves the information of word order in a sentence~-- for the details, please refer to \cite{memn2n}). The same procedure is applied to the user's input using another embedding matrix $B$.

Another important feature in the MemN2N architecture is reading from memory with attention. With the input sentences  and the utterance 
encoded, the match between each of the memory vectors $m_i$ and the utterance $u$ is calculated:

$$p_i = Softmax(u^T m_i)$$

This is used as the attention vector over the encoded memories further in the inference process.

Next, for the final answer prediction, both attention-weighted memories and user's utterance are passed through the final weight matrix $W$:

$$\hat{a} = Softmax(W(o + u))$$

where $o = \sum_i{p_i c_i}$ is weighted memories.

For the QA tasks, the answer $\hat{a}$ is just an index of a word from the vocabulary. In dialogue tasks, however, answers are the entire utterances, either system utterances (e.g.\ \textit{``how many people would be in your party?''}) or API calls (\texttt{``api\_call french london four expensive''}). They are still predicted as indices from the answer candidates list, but given that there is e.g.\ absolutely no overlap in exact api call examples between train and test sets, an internal representation of each candidate answer is added to the architecture \cite{babi}. Thus, the final step now looks as follows:

$$\hat{a} = Softmax((o + u)^T \cdot W(y))$$

where $y$ is a vector of answer candidates processed just as described above for the input sentences, with $W$ as the embedding matrix.

The architecture described above may be stacked into several layers called hops (Figure~\ref{fig:memn2n} (b))~-- refer for details to \cite{memn2n}; here  we're   initially interested in the single hop configuration (see the next section), for which \cite{babi} report their results.

\begin{table*}[ht]
\begin{center}
\begin{tabular}{|c|c|c|c|c|c|}\hline
 train / test set configuration&train accuracy&test accuracy\\\hline \hline 
\textbf{bAbI / bAbI}&100&100\\\hline
\textbf{bAbI / bAbI+}&100&28\\\hline
\textbf{bAbI+ / bAbI}&67&99\\\hline
\textbf{bAbI+ /  bAbI+}&72&53\\\hline
\end{tabular}
\end{center} \caption{MemN2N API call accuracy (\%)}\label{tab:api_call_accuracy}
\end{table*}

\begin{table*}[ht]
\begin{center}
\begin{tabular}{|c|c|c|c|c|}\hline
 training bAbI+ dialogues&memory hops&embedding size&train accuracy&test accuracy\\\hline \hline 
\textbf{2000}&2&128&72.5&57.5\\\hline
\textbf{5000}&2&128&72.7&60.7\\\hline
\textbf{10000}&2&128&72.8&65.8\\\hline
\textbf{50000}&1&128&82.6&78.2\\\hline
\textbf{100000}&1&64&83.3&80.5\\\hline
\end{tabular}
\end{center} \caption{MemN2N API call accuracy (\%) with extended training data}\label{tab:memn2n_extended_data}
\end{table*}

\subsection{Data preprocessing and the \memnn setup}

In order to adapt the data for the MemN2N, we transform the dialogues into \textit{<story, question, answer>} triplets. The number of triplets for a single dialogue is equal to the number of the system's turns, and in each triplet, the {\tt answer} is the current system's turn, the {\tt question} is the user's turn preceding it, and the {\tt story} is a list of all the previous turns from both sides. Other than that, each sentence in the {\tt story} gets 2 additional tokens: the number of the turn, and the ID of the speaker \cite{babi}.

We also use the single embedding matrix $A$ for both input memories and the user's question; the same matrix is used for the output memories representation~-- in that we follow \cite{babi}, and it corresponds to the ``Adjacent'' weight tying model in \cite{memn2n}.

In our setup, there are no out-of-vocabulary words for the model during both training and testing, and for both bAbI and bAbI+ with the maximum sentence length taking account of the increase due to the transformations in bAbI+.

We train our \memnns with a Stochastic Gradient Descent optimizer for \textbf{100} epochs with a learning rate of \textbf{0.01} and a batch size of \textbf{8}~-- in this we again follow the configuration reported by \cite{babi} to be the best for  bAbI  Task 1.




\subsection{Experiments}
We are here interested in: (1) how robust \memnns are to the surface transformations in bAbI+ when trained on bAbI; (2) can \memnns learn to interpret bAbI+ when they are in fact trained on similar data that actually contain the bAbI+ structures~-- i.e. when trained on bAbI+; and (3) if so, how much bAbI+ data is needed for this. While (1) is a question about generalisation properties of a model, (2) \& (3) are about potential in principle and/or practical limitations of \memnns to learn to interpret dialogues containing, e.g. self-corrections where utterances contain both the correct, and an incorrect (and subsequently repaired) slot value (e.g. ``for four sorry five people''). To answer (1) we therefore train the model on the bAbI dataset and test on bAbI+; and to answer (2) \& (3) we train the model on the bAbI+ train set and test it on the bAbI+ test set. Furthermore, in order to explore the impact of the amount of training data on the model's performance, we perform the latter experiment with varying train set size, as well as varying the hyperparameters, embedding size \& number of hops. The extended training data is obtained in the same way as the initial bAbI+ dataset: we go over the same original bAbI dialogues and keep randomly mixing in the incremental modifications. 

\paragraph{Performance Measure: Semantic Accuracy} Self-corrections  and restarts are especially problematic because processing them is potentially a non-monotonic operation involving deletion and replacement in the resulting semantic representations. To measure the model's effectiveness in processing such structures we therefore consider the \textit{semantic accuracy} of the model defined as how accurately it predicts the final API calls~-- recall that the API calls contain all the values of the slots corresponding to the user's request expressed in the preceding dialogue.

\paragraph{Hypotheses} We predicted that (i) given the positional encoding of memory vectors in the \memnn model and the attendant attention mechanisms, it would be able to learn to process bAbI+ dialogues given that it was trained on similar data, resulting in an insignificant drop in performance from bAbI to bAbI+ data; (ii) a lot more data would be needed to learn to process the bAbI+ structures; and (iii) if trained on bAbI data, there would be a very significant drop in performance on bAbI+ with incorrect API calls predicted as a result of incorrect weightings and total lack of opportunity to learn the meaning of words such as ``no'' or ``sorry'' which trigger the self-corrections and restarts.





Finally, we also perform training on bAbI+ and testing on bAbI to see if the model is able to generalise from more complex back to the simpler data.



\subsection{Results and Discussion}

\subsubsection{The original setup} \label{original_setup}

Table~\ref{tab:api_call_accuracy} shows how the \memnn model performs in different conditions. For this, we used identical hyperparameter settings to those of \newcite{babi}: \textbf{1} hop, \textbf{128} embedding size, \textbf{100} epochs, learning rate of \textbf{0.01}, and batch size of \textbf{8}. The train and test sets each contain 1000 dialogues, i.e. the entire corpus.

First note that the first row shows identical results to those of \newcite{babi}: training on bAbI and testing on the bAbI test set results in 100\% accuracy in API call prediction. 
It is therefore highly unlikely that the rest of the results reported here are due to implementational differences between our setup and that of \newcite{babi}.


As we had predicted, the model performs very badly when trained on bAbI and tested on bAbI+ showing very poor robustness to the variations we had introduced, and indicating significant overfitting to the original data.

When the model is trained on bAbI+ data, its performance on the bAbI+ API calls nearly doubles, showing that the model can potentially learn to process the bAbI+ test set given enough data~-- see below. Nevertheless, it remains very low at about 53\% making any system created in this fashion unusable in the face of spontaneous dialogue data. We also note that the accuracy on the train set itself is now lower. This suggests that bAbI+ is a dataset significantly harder to learn (or overfit to), and given the extreme homogeneity of the original bAbI train and test sets, overfitting might be one reason for the model's outstanding results. However, training on bAbI+ and testing on bAbI shows that our assumption about the model's ability to generalize to more simple data appears to be correct.

\subsubsection{How much data is enough data?}

Table~\ref{tab:memn2n_extended_data} shows how \memnn performs on the same initial, fixed bAbI+ test set, when trained on progressively more data and up to 100000 bAbI+ dialogues. As \memnn's performance on bigger data highly depends on the model's hyperparameters, in this experiment we perform a grid search over the number of memory hops (1, 2, 3), and the embeddings dimensionality (32, 64, 128) for each train set size independently~-- everything else is fixed as in the previous experiment.
The table only shows the best performing hyperparameter configuration for each of the train set sizes.

The results confirm hypothesis (ii) above, i.e. that \memnns are in principle able to learn to process the incremental dialogue phenomena in bAbI+ but that they require tens of thousands of training instances for this: even with 100000 dialogues, the semantic accuracy on the original test set stands at 80.5\%.


These experiments shed significant light on the currently ambiguous robustness results reported in the dialogue systems literature today. Specifically, they show that, from the point of view of dialogue system developers in the real world, learning to process natural spontaneous dialogue using \emph{\memnns only} in an end-to-end fashion may not be practical: in bAbI+, the disfluent incremental phenomena were mixed in at will, thus affording access to arbitrarily large training sets; furthermore, the test set was synthetically constructed to follow the same pattern as in the train set; whereas real, natural, spontaneous dialogue data is not only very expensive to collect, but is bound to be more complex, with the closeness between train \& test data very difficult to control.

A potential solution to this `small data' problem is the use of computational dialogue models (such as e.g. \cite{Ginzburg12,Larsson02,Poesio.Rieser10,Eshghi.etal15}) with studied empirical foundation as a form of bias or prior in subsequent learning, thus exploiting the linguistic knowledge inherent in such models. Even if they are not used directly, they can be used to inform the architecture of particular machine learning methods, especially deep learning architectures and techniques, with a view to more modularity in such architectures, with general language processing modules that are transferable from one domain to another, much like a NL grammar.




\section{Testing an incremental, semantic grammar on bAbI \& bAbI+}\label{dylan}
\begin{figure*}[!ht]
\begin{footnotesize}
\centerline{
\begin{tabular}{c@{$\quad\mapsto\quad$}c@{$\quad\mapsto\quad$}c@{$\quad\mapsto\quad$}c}
$\ttrnode{}{event&e_s\\p1_{=today(event)}&t}$ &
$\ttrnode{}{event_{=arrive}&e_s\\p1_{=today(event)}&t\\p2_{=pres(event)}&t\\x_{=robin}&e\\p3_{=subj(event,x)}&t}$ &
$\ttrnode{}{event_{=arrive}&e_s\\p1_{=today(event)}&t\\p2_{=pres(event)}&t\\x_{=robin}&e\\p3_{=subj(event,x)}&t\\x1&e\\p3_{=from(event,x1)}&t}$ &
$\ttrnode{}{event_{=arrive}&e_s\\p1_{=today(event)}&t\\p2_{=pres(event)}&t\\x_{=robin}&e\\p_{=subj(event,x)}&t\\x1_{=Sweden}&e\\p3_{=from(event,x1)}&t}$
\\[5em]
\emph{``A: Today''} & \emph{``..Robin arrives''} & \emph{``B: from?''} & \emph{``A: Sweden''}
\end{tabular}
}
\end{footnotesize}
\caption{Incremental parsing with DyLan}\label{fig:subtype}

\end{figure*}

\begin{figure*}
\includegraphics[width=\linewidth]{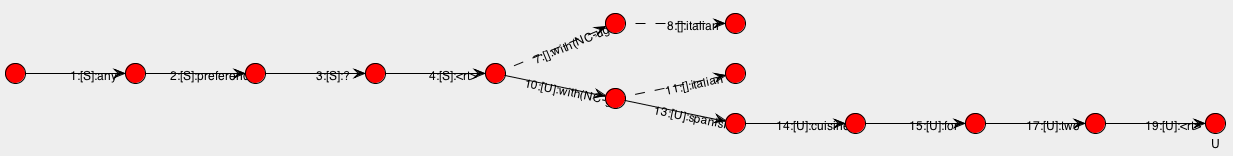}
\caption{Processing self-corrections \& restarts with DyLan: ``A: any preference? B: with italian yeah sorry with spanish cuisine''}\label{fig:dag}
\end{figure*}

In this section, we first quickly introduce an incremental, semantic parser for dialogue processing~-- DyLan \cite{Eshghi.etal11,Eshghi15,Purver.etal11}~-- based around the Dynamic Syntax and Type Theory with Records framework \cite{Kempson.etal01,Cann.etal05a,Eshghi.etal12,Cooper05,Cooper12}, which has been used recently in combination with Reinforcement Learning for automatically inducing fully incremental dialogue systems from small amounts of raw, unannotated dialogue data \cite{Eshghi.Lemon14,Kalatzis.etal16}, showing remarkable generalisation properties \cite{Eshghi.etal17,Eshghi.etal17b}. We then go on to perform the same experiments on semantic accuracy as we did above with \memnns using this linguistically informed model instead.

\subsection{DyLan\footnote{DyLan is derived from ``Dynamics of Language''}: parser for Dynamic Syntax} 

DyLan \cite{Eshghi.etal11,Eshghi15} is the parser/implementation for Dynamic Syntax (DS), an action-based, word-by-word incremental, semantic grammar formalism \cite{Kempson.etal01,Cann.etal05a}, especially suited  to the highly fragmentary and contextual nature of dialogue. In DS, words are conditional actions~-- semantic updates; and dialogue is modelled as the interactive and incremental construction of contextual and semantic representations \cite{Eshghi.etal15}~-- see Fig.~\ref{fig:subtype} which shows how semantic representations are constructed incrementally as Record Types of Type Theory with Records (TTR) \cite{Cooper05,Cooper12}. The contextual representations afforded by DS are of the fine-grained semantic content that is jointly negotiated/agreed upon by the interlocutors, as a result of processing questions and answers, clarification interaction, acceptances, self-/other-corrections, restarts, and other characteristic incremental phenomena in dialogue~-- see Fig.~\ref{fig:dag} for a sketch of how self-corrections and restarts are processed via a backtrack and search mechanism over the parse search graph. The nodes in this graph are (partial) semantic trees, and the edges correspond to words uttered by particular speakers. Context of a partial tree in DS is the path back to root on this parse search graph (see \newcite{Hough15,Hough.Purver14,Eshghi.etal15} for details of the model). 
The upshot of this is that using DS, one can not only track the semantic content of some current turn as it is being constructed (parsed or generated) word-by-word, but also the context of the conversation as whole, with the latter also encoding the grounded/agreed content of the conversation (see Eshghi et al. (2015); Purver et al. \shortcite{Purver.etal10} for details). Crucially for \cite{Eshghi.etal17}'s model, the inherent incrementality of DS together with the word-level, as well as cross-turn, parsing constraints it provides, enables the word-by-word exploration of the space of grammatical dialogues, thus lending itself very well to Reinforcement Learning \cite{Kalatzis.etal16,Eshghi.etal17b}.

\subsection{Parsing bAbI and bAbI+ dialogues with DS}
The Dynamic Syntax (DS) grammar is learnable from data \cite{Eshghi.etal13a,Eshghi.etal13b}. But since the lexicon was induced from a corpus of child-directed utterances in this prior work, there were some constructions as well as individual words that it did not include\footnote{in the near future we will use the learning method in \newcite{Eshghi.etal13a} to induce DS grammars from larger semantic corpora such as the Groningen Meaning Bank, leading to much more wide-coverage lexicons than the present one}. One of the authors therefore extended this induced grammar manually to cover the bAbI dataset, which, despite not being very diverse, contains a wide range of complex grammatical constructions, such as long sequences of prepositional phrases, adjuncts, short answers to yes/no and wh-questions, appositions of NPs, causative verbs etc --  and all of this within and across dialogue turns/speakers.


Using DyLan we parsed all dialogues in the bAbI train and test sets, as well as on the bAbI+ corpus word-by-word, including both user and system utterances, in context. The grammar  parses 100\% of the dialogues, i.e.\ it does not fail on any word in any of the dialogues.

\subsection{Semantic Accuracy of DyLan}
 Merely parsing all dialogues in the bAbI and bAbI+ datasets doesn't mean that the semantic representations compiled for the dialogues were in fact correct. To measure the semantic accuracy of the parser, we used, as before, the API call annotations at the end of bAbI and bAbI+ task 1 dialogues. This was done programmatically by checking that the correct slot values~-- those in the API call annotations~-- were in fact present in the semantic representations produced by the parser for each dialogue (see Fig.~\ref{fig:subtype} for example semantic representations). We further checked that there is no other incorrect slot value present in these representations.

The results showed that the parser has 100\% semantic accuracy on both bAbI and bAbI+. This result is not surprising, given that Dynamic Syntax is a general model of incremental language processing, including phenomena such as self-corrections \& restarts (see \cite{Hough15} for details of the model)\footnote{A helpful reviewer points out that the DyLan setup is a carefully tuned rule-based system, thus rendering these results trivial. But we note that the results here are not due to ad-hoc constructions of rules/lexicons, but due to the generality of the grammar model, and its attendant incremental, left-to-right properties. For example, the ability to process self-corrections, restarts, etc. ``comes for free'', without the need to add or posit new machinery}. It is worth noting that even though new lexical entries would have to be added for each new dataset/domain, given the parts-of-speech of the words in any given dataset, this can mostly be done automatically. 

Moreover, this result further reinforces the point made by \newcite{Eshghi.etal17b}
 about the generalisation power of the Dynamic Syntax grammar: the grammar automatically generalises to a combinatorially large number of dialogue variations with various phenomena such as self-corrections, hesitations, restarts, clarification interaction, continuations, question-answer pairs etc. without having actually observed these in any of the seed/training dialogues.

\section{Conclusion and ongoing work}

Our main advance is in exploring  incremental processing for wider coverage of more natural everyday dialogue (e.g.\ containing self-corrections). 


Our experiments show that a state-of-the-art model for end-to-end goal-oriented dialogue, \memnn, lacks the ability to generalise to such phenomena, and performs poorly when confronted with natural spontaneous dialogue data. Our experiments further show that although this particular model is in principle able to learn to process incremental dialogue phenomena, it requires an impractically large amount of data to do so. The results in this paper therefore shed significant light on the currently ambiguous robustness results reported for end-to-end systems.


We also assessed the performance of the DyLan dialogue parser on bAbI and  bAbI+ which showed 100\% parsing and semantic accuracy, highlighting the generalisation power of models that are linguistically informed, and theoretically grounded as compared with pure machine learning methods that aim to learn to process dialogue bottom up from textual data alone, without any linguistic bias. These issues are explored further in \cite{Eshghi.etal17b}.


\bibliography{babble,all,SG_B}
\bibliographystyle{eacl2017}

 \end{document}